\documentclass[journal]{IEEEtran}
\ifCLASSINFOpdf
\else
\fi
\usepackage{times}
\usepackage{epsfig}
\usepackage{graphicx}
\usepackage{amsmath}
\usepackage{amssymb}
\usepackage{graphicx}
\usepackage{subfigure}
\usepackage{amsmath}
\usepackage{multirow}
\usepackage{bm}

\usepackage[pagebackref=true,breaklinks=true,letterpaper=true,colorlinks,bookmarks=false]{hyperref}
\begin{document}
%
\title{When Work Matters: Transforming Classical Network Structures to Graph CNN}
%
%
%
%

\author{
	Wenting Zhao$^1$,
	Chunyan Xu$^2$,
	Zhen Cui$^3$,
	Tong Zhang$^4$,
	Jiatao Jiang$^5$,
	Zhenyu Zhang$^6$,
	Jian Yang$^7$
	\\
	$^1$ $^2$ $^3$ $^5$ $^6$ $^7$ Nanjing University of Science and Technology
    \\
	$^4$ Southeast University
	%

\thanks{Email address: \{wtingzhao, cyx, zhen.cui, csjyang\}@njust.edu.cn (W. Zhao, C. Xu, Z. Cui and J. Yang), tongzhang@seu.edu.cn (T. Zhang), zhangjesse@foxmail.com (Z. Zhang).}

}

\maketitle

\begin{abstract}
%
Numerous pattern recognition applications can be formed as learning from graph-structured data, including social network, protein-interaction network, the world wide web data, knowledge graph, etc.
While convolutional neural network (CNN) facilitates great advances in gridded image/video understanding tasks, very limited attention has been devoted to transform these successful network structures (including Inception net, Residual net, Dense net, etc.) to establish convolutional networks on graph, due to its irregularity and complexity geometric topologies (unordered vertices, unfixed number of adjacent edges/vertices).
In this paper, we aim to give a comprehensive analysis of when work matters by transforming different classical network structures to graph CNN, particularly in the basic graph recognition problem.
Specifically, we firstly review the general graph CNN methods, especially in its spectral filtering operation on the irregular graph data.
%
We then introduce the basic structures of ResNet, Inception and DenseNet into graph CNN and construct these network structures on graph, named as G\underline{\hspace{0.5em}}ResNet, G\underline{\hspace{0.5em}}Inception, G\underline{\hspace{0.5em}}DenseNet.
In particular, it seeks to help graph CNNs by shedding light on how these classical network structures work and providing guidelines for choosing appropriate graph network frameworks.
Finally, we comprehensively evaluate the performance of these different network structures on several public graph datasets (including social networks and bioinformatic datasets), and demonstrate how different network structures work on graph CNN in the graph recognition task.


\end{abstract}


%
\IEEEpeerreviewmaketitle
\section{Introduction}
A graph-structured data sample consists of a finite set of vertices/nodes, together with a set of connections revealing the relationship between unordered pairs of these vertices (named edges).
Numerous important high-level applications, especially in the increasingly connected and blended world, can be framed as learning from graph data, including social network~\cite{graphdata_associate,saen2017}, protein-interaction network~\cite{proteininteraction}, the world wide web data~\cite{www}, knowledge graph, etc.
Among these practical problems, learning an appropriate neural network from such structured graphs becomes the most critical topic.

Recently, convolutional neural networks (CNNs), together with multiple evolved variants, have achieved very promising performance in processing grid-shaped images/videos~\cite{alexnet,vgg,densenet,resnet}, such as image recognition, object detection, depth estimation, image restoration, object segmentation, etc.
%
LeCun $\textit{et al.}$~\cite{Lecun98} originally trained the layer-wise convolutional network with the back-propagation algorithm in 1998.
Krizhevsky $\textit{et al.}$~\cite{alexnet} introduced a similar CNN network with a deeper and much wider version, and achieved a breakthrough, outperforming the existing handcrafted features on ILSVRC 2012 competition.
GoogLeNet~\cite{googleLeNet} proposed more effective inception module to design a local network topology by adopting multiple receptive field sizes.
VGGNet~\cite{vgg}, which consisted of 16 convolutional layers, was also very appealing because of its very uniform architecture.
He $\textit{et al.}$~\cite{resnet} introduced a substantially deeper architecture (dubbed Residual Neural Network, ResNet)) with “skip connections”, which are also known as gated units or gated recurrent units.
Dense Convolutional Network (DenseNet)~\cite{densenet} further proposed a densely connected structure, which can connect each layer to every other layer in a feed-forward fashion.
%
However, due to the irregularity and complexity geometric topologies of graph-structured data, these successful CNN structures on representing grid-shaped image/video data cannot be straightforwardly applied to the graph data, especially these elementary operators including convolutional filtering, pooling, translation, etc.

Driven by the developments and limitation of above CNNs,  various algorithms devoted to graph CNNs have been proposed in previous literatures.
In general, these algorithms can be divided into two main categories according to their ways of conducting convolution on graphs: one can be named as spatial graph CNNs convolving directly on graphs according to the spatial information of nodes,  while  the other employs spectral filtering based on Spectral Graph Theory~\cite{chung1997spectral}.
The former one is  analogues to  CNN which firstly constructs a window of certain size for nodes on the graph and then  conducts convolution operation~\cite{niepert2016learning}. However,
its disadvantage is the loss of structural information of graph data. To overcome this shortcoming, DGCNN \cite{DGCNN} proposed dynamic convolutional kernel to adapt to the size and order of the node neighborhood, supporting different scales of convolutional receptive fields. As another kind of spatial graph CNNs, a mixture model networks (MoNet) is proposed in \cite{monet}, using a parametric patch to match template for convolution operations on graphs or manifolds. Moreover,  as a particular instance of MoNet~\cite{monet}, \cite{2017GraphAttentionnNet}
introduces an attention-based architecture to perform node classification of graph-structured data. The latter category of graph CNNs is based on the Spectral Graph Theory \cite{chung1997spectral}. In the graph setting, the graph Laplacian eigenvalues and eigenvectors provide a notion of frequency \cite{shuman2013emerging} and spectral filtering has been successfully applied to the field of graphs. In \cite{BrunaZSL13}, CNN was firstly generalized to graph using spectral filtering. On the basis, \cite{henaff2015deep} considered large-scale classification problems with small learning complexity and further made some extension.~\cite{defferrard2016conv} approximated smooth filters in the spectral domain using Chebyshev polynomials with free parameters that are learned in a neural network-like model. To further improve computation efficiency,~\cite{kipf2016semi} presented a first-order approximation of spectral graph convolutions, and in many cases allowed both for significantly faster training times and higher predictive accuracy.
Although these approaches have devoted to design more effective convolutional filtering and pooling on graph, very limited attention has been focused on how to transform these successful network structures (including Inception net, Residual net, Dense net, etc.) to establish more promising Graph CNN architectures on graph domains.





In this paper, we pay attention to give a comprehensive analysis of when work matters by transforming different classical network structures to graph CNN, particularly in the basic graph recognition problem.
Specifically, we firstly review the general graph CNN methods, especially in its spectral filtering operation on the irregular graph data.
%
We then introduce the basic structures of ResNet, Inception and DenseNet into graph CNN and extend these network structures on graph domains, named as G\underline{\hspace{0.5em}}ResNet, G\underline{\hspace{0.5em}}Inception, G\underline{\hspace{0.5em}}DenseNet.
G\underline{\hspace{0.5em}}ResNet with a similar structure with ResNet~\cite{resnet}, is a multi-layer gated graph CNN by learning residual functions with reference to the layer inputs.
G\underline{\hspace{0.5em}}Inception can better consider the different receptive fields of graph signal and also increase the width of the network while keeping the computational budget constant.
G\underline{\hspace{0.5em}}DenseNet can strengthen feature propagation and encourage graph signals of all preceding layers in the graph CNN.
In particular, it seeks to help graph CNNs by shedding light on how these classical network structures work and providing guidelines for choosing appropriate graph network frameworks.
We comprehensively evaluate the performance of these different network structures on several public graph datasets (including social networks and bioinformatic datasets), and demonstrate how different network structures work on graph CNN in the graph recognition task.

The remainder of this paper is organized as follows. Section \ref{sec:relatedwork} reviews some related works about classical network structures and methods of graph CNN. Section \ref{sec:Convolution on graph} briefly introduces the approach we used to convolve on the graph in this paper. Section \ref{sec:StruToGraCNN} presents the three network structures of G\underline{\hspace{0.5em}}ResNet, G\underline{\hspace{0.5em}}Inception and G\underline{\hspace{0.5em}}DenseNet. Section  \ref{sec:exp} describes the implementation details, reports the performance of our three graph CNN structures on social networks and bioinformatic datasets and makes some discussions. Finally, Section  \ref{sec:con} concludes this paper and gives future research direction.

\section{Related work}
\label{sec:relatedwork}
\textbf{Classical network structures:}
Deep network learning has been long studied for dealing with these gridded image/video understanding problems~\cite{pr2017facial, pr2016survey, simonyan2014two, pr2018dvideo}. In 1998, LeCun $\textit{et al.}$~\cite{Lecun98} trained multilayer neural networks with the back-propagation algorithm and the gradient learning technique, and then demonstrated its effectiveness on the handwritten digit recognition task. 
Recently, for further boosting its discriminative capability, there has been a resurgence of research interest in the exploration of various network structures.
AlexNet~\cite{alexnet} is a special type of deep CNN model and achieves a breakthrough, outperforming the existing handcrafted features on ILSVRC 2012 which contains 1000 object classes.
Another deep network structure, namely ``Network In Network" (NIN)~\cite{NIN} is proposed to build a micro network with more complex structures to abstract the data within the receptive field, and the proposed $1 \times 1$ convolution kernel is later applied in the GoogLeNet model to reduce the dimension and avoid computational explosion.
VGGNet~\cite{vgg} consists of 16 convolutional layers and is very appealing because of its very uniform architecture. And It is currently the most preferred choice in the community for extracting features from vision inputs.
%
A more effective inception module, introduced by GoogLeNet~\cite{googleLeNet} model, can be employed to design a local network topology. It convolves with different sized kernels, concatenates the results as input to the next layer, and implements the use of multi-scale features.
%
ResNet \cite{resnet} can easily improve performance by significantly increasing the depth of network compared to the “plain” nets (that simply stack layers). And it addresses the degradation problem \cite{7299173, HighwayNetworks} of accuracy  that arises with the network depth increasing by introducing a deep residual learning framework.
%
%
Through the connection between each convolution layer, DenseNet \cite{densenet} encourages feature reuse and then learns more and compact features while keeping fewer parameters and less computation by using $1 \times 1$ convolution.
%


\textbf{Graph CNN:}
Recent CNN on graph data has raised a progressive direction in the problem of graph recognition. At present, there are two main categories to execute the convolution operation on graphs. One is analogous to the common CNN in the gridded image/video samples, constructing a window of a certain size on the graph, and then doing the convolution by a fixed-size template. The other is based on Spectral Graph Theory\cite{chung1997spectral}. Features of the graph are first transformed into the frequency domain using the Fourier transform and convolved on the spectrum of the graph. Then  an inverse Fourier transform is done to transform the feature maps to vertex domain.

Based on image CNN, the work of \cite{niepert2016learning} proposed a general method to learn representation for arbitrary graphs and transformed the data of a graph structure into a structure that CNN can efficiently handle. It mainly consists of two steps: selecting a representative node sequence from the graph structure and finding a convolutional neighborhood for each selected node. However, in this process, structural information of the graph may be lost and also some redundant information can be introduced. Therefore, the paper \cite{DGCNN} introduced a Gaussian mixture model by adding a disordered graph convolutional layer (DGCL) to overcome the above problems. Besides, ~\cite{2017GraphAttentionnNet} presented masked self-attentional layers to adptively endow different weights to those neighbor vertices of a vertex, and thereby the performance of the model can be improved.  To  capture temporal evolution of graph sequences,  recursive neural networks are introduced into the graph in \cite{seo2016recurrent}.

Based on the Spectral Graph Theory\cite{chung1997spectral}, a generic framework for processing data on graphs is presented in \cite{shuman2013emerging}, and the fundamental operations such as filtering, translation, modulation, dilation, and downsampling are generalized to the graph setting.
\cite{BrunaZSL13} proposed two efficient constructions: Spatial Construction and Spectral Construction. Then, \cite{henaff2015deep} extended \cite{BrunaZSL13} to large-scale classification problems with small learning complexity, and proposed unsupervised and new supervised graph estimation strategies. In \cite{defferrard2016conv}, the spectral graph theoretical of CNNs on graphs is formulated and strictly localized  spectral filters are given. More importantly, recursive Chebyshev polynomials are utilized to approximate parameterized polynomial filters such that the complexity and efficiency of learning are significantly reduced.~\cite{kipf2016semi} was based on spectral graph convolutional neural networks \cite{defferrard2016conv, BrunaZSL13} simplifying to a linear function of first-order and conducted the task of transductive node classification in a large-scale network.
The paper \cite{AGCNN} learned a residual Laplacian matrix for each graph and the optimal distance metric parameters shared among the data, then graphs of arbitrary structure and size can be input into the CNN.


\begin{figure*}[ht]
	\centering
	\subfigure[]{\includegraphics[scale=0.9]{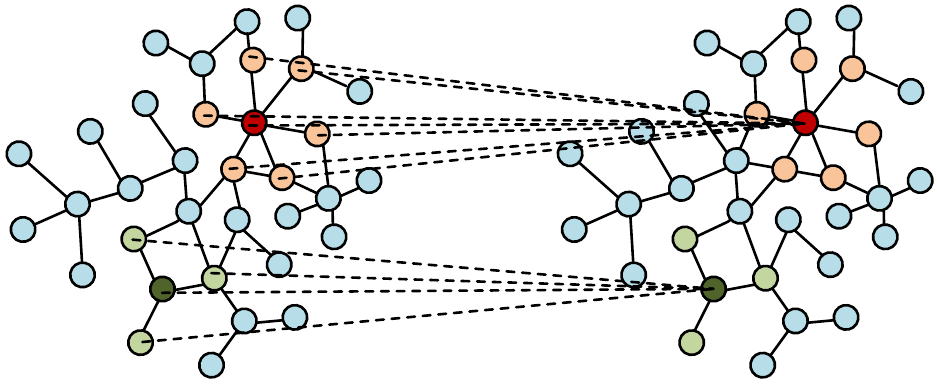}}
	\subfigure[]{\includegraphics[scale=1]{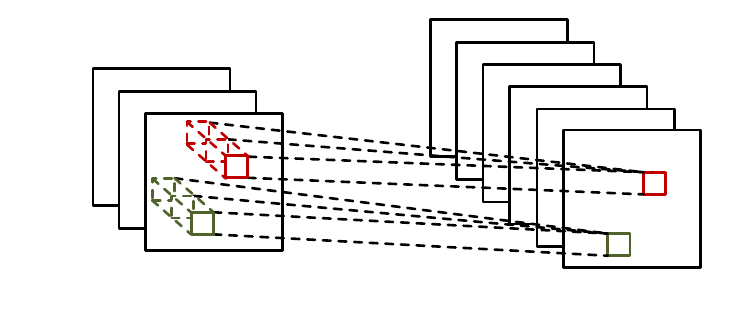} \vspace{10em}}
	\vspace{-1em}
	\caption{Different convolution operations on graph (left) and gridded image (right) samples.
	(a) shows the 1-localized convolution on graph data, where the red node has 6 adjacencies, and the green node has 3 adjacencies. In the process of convolution, they use the features of the 6 and 3 nodes respectively, in addition to the feature of their own nodes. 	
	(b) shows the convolution operation on the regular gridded data (e.g., images and videos). As long as a fixed-size convolution kernel is given, the convolution kernel will slid from left-to-right and top-to-bottom, all vertices can be then convolved once.}
	\label{graph-image-conv}
	\vspace{-1em}
\end{figure*}

\section{Convolution on graph}
\label{sec:Convolution on graph}

As we have introduced, the graph-structured data is with the irregular structure and completely coordinate-free on vertices and edges.
To generalize the idea of common CNNs onto graphs, we would like to review the convolution filter on homomorphic graphs/subgraphs, which is different to the convolution on these gridded images/videos, as shown in Fig.~\ref{graph-image-conv}.
Here we mainly introduce the Spectral filtering of graphs, which is also used in our followed experiments.

For a graph $\mathcal{G} \left(\mathcal{V},\mathcal{E} \right)$, $\mathcal{V}$ represents a set of vertices with the number $|\mathcal{V} |= n $ and $\mathcal{E}$ represents a set of edges.  Let $\mathbf{W}$ denotes the adjacent matrix representing the topology of $\mathcal{G}$, then $W_{ij} = 1$ if  there is an edge connection between vertices $v_i$ and $v_j$,  otherwise $W_{ij}$ = 0. Each vertex has a feature or signal denoted as  $\mathbf{x}\in\mathbf{R}^{d}$, and $d$ is the number of feature. The features of all vertices in graph are summarized by $\mathbf{X}\in\mathbf{R}^{{n}\times{d}}$.


Laplacian matrix of combinational definition is $\mathbf{L=D-W}\in\mathbf{R}^{n \times n}$, where $\mathbf{D} \in \mathbf{R}^{n \times n}$ is the degree matrix with $D_{ij} = \sum_{j}W_{ij}$, and the normalized definition is $\mathbf{L=I}_{n}-\mathbf{D}^{{-1/2}}\mathbf{W} \mathbf{D}^{-1/2}$, where $\mathbf{I}_{n}$ is identify matrix. As Laplacian matrix is positive semidefinite, there exists a matrix denoted as $\mathbf{U}$ which is composed of a set of orthogonal eigenvectors $\left\{ {\mathbf{u}_1,\mathbf{u}_2,\mathbf{u}_3 \cdots \mathbf{u}_n} \right\}$ satisfying  $\mathbf{L = U \Lambda} \mathbf{U}^T$, where $\mathbf{\Lambda}$ is a diagonal matrix consisting of eigenvalues  $\left\{ {\lambda_1,\lambda_2,\lambda_3 \cdots \lambda_n} \right\}$.

In Fourier domain, the eigenvalue represents a specific frequency. The eigenvectors $\left\{ \mathbf{u}_1,\mathbf{u}_2,\mathbf{u}_3 \cdots \mathbf{u}_n \right\}$ are Fourier basis. Then, the graph Fourier transform can be formulated as
\begin{equation}
\mathbf{\hat{X}=U}^T\mathbf{X}
\end{equation}
and the inverse transformation is given by
\begin{equation}
\mathbf{X=U\hat{X}}
\end{equation}

In the Fourier domain, the filtering operation of graph is given as
\begin{equation}
\mathbf{\hat{X}}_{out}=  \hat{h} (\mathbf{\Lambda}) \mathbf{\hat{X}}
\end{equation}
where $\hat{h}(\cdot)$ is the spectral operator.
Equivalently, we transform it to the time domain as
\begin{equation}
\begin{aligned}
\mathbf{X}_{out} &= \mathbf{U} \hat{h} \mathbf{(\Lambda )} \mathbf{\hat{X}} \\
&= \mathbf{U} \hat{h} \mathbf{(\Lambda)} \mathbf{U}^T \mathbf{X}  \\
&= \mathbf{U} \left(\begin{matrix}
\hat{h}( \lambda_{1}) & \\
& \hat{h}( \lambda_{2}) \\
& & \ddots \\
& & & \hat{h}(\lambda_{n})
\end{matrix}\right) \mathbf{U}^T \mathbf{X} \\
&= \hat{h}(\mathbf{L}) \mathbf{X}
\end{aligned}
\end{equation}

The spectral filtering is parameterized as
\begin{equation}
\begin{aligned}
\hat{h}_\theta \left( \mathbf{\Lambda} \right)
& = \left(\begin{matrix}
\hat{h}_{\theta}  \left( \lambda_{1} \right) & \\
&  \hat{h}_\theta \left( \lambda_{2} \right) \\
& & \ddots \\
& & & \hat{h}_\theta \left( \lambda_{n} \right)
\end{matrix}\right) \\
&= \sum_{k=0} \theta_k \mathbf{\Lambda}^k
\end{aligned}
\end{equation}


In order to reduce the complexity of learning, Chebyshev are used to approximate as
\begin{equation}
\hat{h}_{\theta^{'},k} (\mathbf{\Lambda})
= \sum_{k=0}^K {\theta^{'}_k} T_k ( \mathbf{\Lambda} )
\end{equation}
where $\theta^{'}_k$ is the Chebyshev polynomial coefficient, $T_k ( \mathbf{\Lambda})$ is the Chebyshev approximate polynomial, $T_0 (\alpha) = 1$, $T_1 (\alpha) = \alpha$ and the recursive formulation is $T_k (\alpha) = 2 \alpha T_{k-1} (\alpha) - 2 \alpha T_{k-2} (\alpha)$.

Then, the Chebyshev approximate local filter is finally given as
\begin{equation}
\mathbf{X}_{out} = \mathbf{U} \hat{h}_{\theta^{'},k}( \mathbf{\Lambda}) \mathbf{U}^T \mathbf{X}
=\sum_{k=0}^K {\theta^{'}_k} T_k ( \mathbf{\tilde{L}} )\mathbf{X}
\label{cheby}
\end{equation}
where $ \mathbf{\tilde{L}} = \frac{2\mathbf{L}}{\lambda_{max}}-\mathbf{I}_{n} $, $\mathbf{I}_{n}$ is identify matrix. In the following experiments, we set $\lambda_{max}=2$.  $\theta^{'}_k$ is the Chebyshev polynomial coefficient that needs to learn.

In the training process, the cross-entropy loss function is employed for optimizing the parameters of graph CNNs.
\begin{equation}
J(\theta)= - \sum_{i}I(y= C_{i}) ln \{\hat{y_i}(\theta) \}
\end{equation}
Samples can be divided into $C$ classes and $ i\in C$. $y$ is the label of the sample. $I(y = C_{i})$ is the indicative function,  $I(y = C_{i})=1$ when $y = C_{i}$, otherwise $I(y = C_{i})=0$.
The back-propagation process can be given as
\begin{equation}
\frac{\partial J(\theta)}{\partial \theta} = - \sum_{i} I(y = C_{i} ) \frac{1}{\hat{y_i}(\theta)} \frac{\partial \hat{y_i}(\theta)}{\partial \theta}
\end{equation}


\begin{figure}[t]
	\centering
	\subfigure[Our baseline]{\includegraphics[scale=0.793]{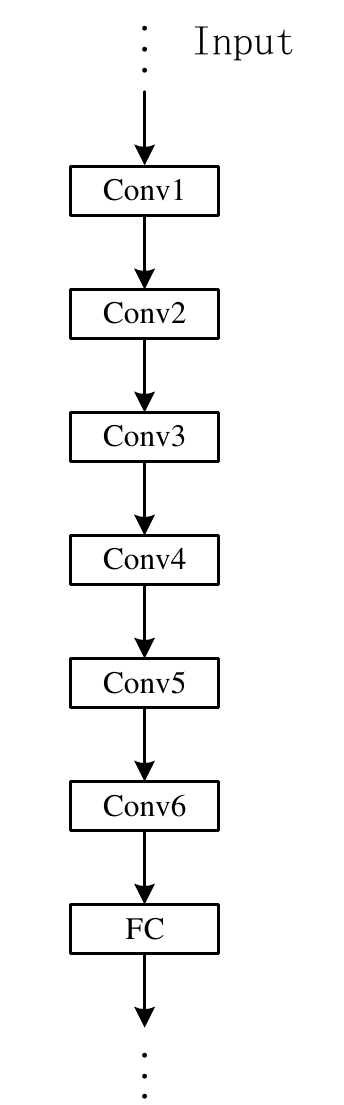} \label{base}}
	\subfigure[G\underline{\hspace{0.5em}}Densenet]{\includegraphics[scale=0.793]{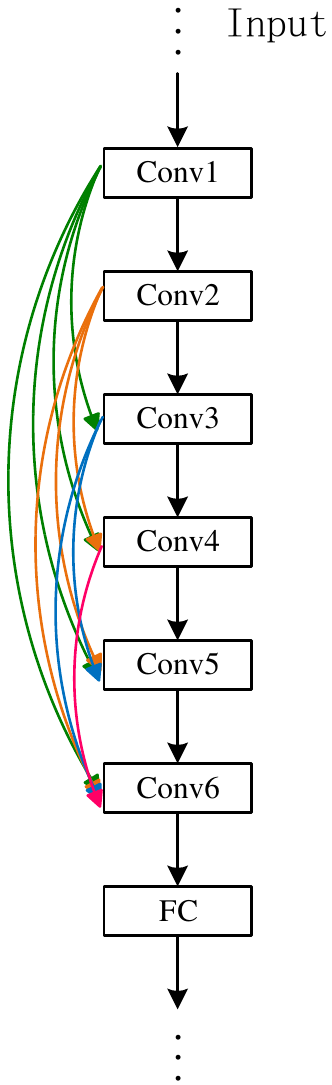} \label{dens}}
	\vspace{-1em}
	\caption{Graph CNN architectures of the baseline and G\underline{\hspace{0.5em}}Densenet.}
	\label{sturcture2}
\end{figure}

\section{Transforming Classic network Structures to Graph CNN}
\label{sec:StruToGraCNN}
In this section, we transform these classic network structures (including ResNet \cite{resnet}, Inception \cite{googleLeNet} and DenseNet \cite{densenet}) to graph CNN and show the structures of  three different networks. We will detailedly introduce four graph CNNs with different network structures, i.e., plain Graph CNN, G\_ResNet, G\_Inception, G\_DenseNet.

%
%
\subsection{Plain Graph CNN}
The plain baseline of graph CNN are mainly inspired by the philosophy of VGG nets~\cite{vgg}, as illustrated in Fig. \ref{base}.
We stack 6 convolution layers, each of which is followed with a batch normalization and rectified linear layer.
Different from the gridded images/videos, on which local convolution kernels can be defined as multiple lattices with various receptive fields, graph CNN can adopt a $K$-localized convolution for every vertex in a graph \cite{defferrard2016conv}. i.e. $K$ hops from the central vertex.
The number of channels in the first three convolution layers is 32, and 64 channels have been set for the last three convolution layers.
According to Eq. \ref{cheby}, in the forward propagation of Graph CNN, we denote $\mathbf{\bar{X}}_k = T_k ( \mathbf{\tilde{L}})\mathbf{X}$, and the $k$-th iteration $\mathbf{\bar{X}}_k = 2 \mathbf{\tilde{L}\bar{X}}_{k-1} - \mathbf{\bar{X}}_{k-2}$ with $\mathbf{\bar{X}}_0 = \mathbf{X}$ and $\mathbf{\bar{X}}_1 = \mathbf{\tilde{L}X}$, where $\mathbf{X}\in\mathbf{R}^{{n}\times{d}}$ is the initial feature map of graph data. For the $K$-localized convolution, $\mathbf{\bar{X}}_K$ is the result of $K$-th iteration.
Here the receptive field $K$ of all convolution layers is simply set to $6$.
The plain Graph CNN ends with a fully-connected layer with softmax, which dimension equals to the number of classes in the graph datasets.

\subsection{Dense Graph CNN}
DenseNet combines features by concatenating them instead of combining features through summation. DenseNet improve flow of information and gradients throughout the network, which makes network easy to train.
Based on the plain Graph CNN, we construct Dense graph CNN (G\underline{\hspace{0.5em}}DenseNet) by connecting each layer to every other layer in a feed-forward fashion, which is shown in Fig. \ref{dens}.
%
%
The input of each convolutional layer is the output of all previous convolutions, and the output of each convolutional layer must be used as the input of following convolutions.
As the number of feature maps per layer gradually increases, the size of $\mathbf{\Theta}_{i}$ also increases from layer to layer. We can formulate Dense graph CNN as
\begin{equation}
\begin{aligned}
& \mathbf{Y}_{0} =\mathcal{F}_{\text{dens}}(\mathbf{\bar{X}}_K, \mathbf{\Theta}_{0})\\
& \mathbf{Y}_{1} =\mathcal{F}_{\text{dens}}(\mathbf{\bar{X}}_K || \mathbf{\bar{Y}}{_{0}}_K, \mathbf{\Theta}_{1})\\
& \mathbf{Y}_{2} =\mathcal{F}_{\text{dens}}(\mathbf{\bar{X}}_K || \mathbf{\bar{Y}}{_{0}}_K || \mathbf{\bar{Y}}{_{1}}_K, \mathbf{\Theta}_{2})\\
&\cdots\\
& \mathbf{Y}_{l} =\mathcal{F}_{\text{dens}}(\mathbf{\bar{X}}_K || \mathbf{\bar{Y}}{_{0}}_K || \cdots || \mathbf{\bar{Y}}{_{l-1}}_K, \mathbf{\Theta}_{l})\\
\label{des}
\end{aligned}
\end{equation}
where $\mathbf{Y}_{0}$, $\mathbf{Y}_{1}$, $\cdots$, $\mathbf{Y}_{l}$ denote the output of different convolution layers, $\mathbf{\bar{X}}_K$, $\mathbf{\bar{Y}}{_{0}}_K$, $\cdots$, $\mathbf{\bar{Y}}{_{l-1}}_K$ correspond to the $K$-localized convolution of input graph signals.
Based on the feature information of all preceding layers, the function $\mathcal{F}_{\text{dnes}}(\cdot)$ represents the dense operation to be learned by convolution layers.



\subsection{Residual Graph CNN }
Based on the above plain network, similar to the ResNet~\cite{resnet}, we insert shortcuts connections which turn the plain Graph CNN into its residual version.
The residual graph CNN is built by stacking two residual graph blocks, each of which consists of the residual part and identity mapping.
The residual part consists of three convolutions, each of which is followed by a batch normalization and a rectified linear layer, and the identity mapping has a linear projection to match the same dimensions.
%
%
The residual graph block is formulated as
%
\begin{equation}
\mathbf{Y} = \mathcal{F}_{\text{res}}(\mathbf{\bar{X}}_K, \{\mathbf{\Theta}_{0}, \mathbf{\Theta}_{1}, \mathbf{\Theta}_{2}\})
+
\mathbf{\bar{X}}_K\mathbf{\Theta}_{s},
\label{res}
\end{equation}
where $\mathbf{\Theta}_{0}$, $\mathbf{\Theta}_{1}$, $\mathbf{\Theta}_{2}$ are the parameters of three different convolution layers, respectively. $\Theta_{s}$ is a linear projection matrix of identity mapping. The function $\mathcal{F}_{\text{res}}(\mathbf{\bar{X}}_K, \{\mathbf{\Theta}_{0}, \mathbf{\Theta}_{1}, \mathbf{\Theta}_{2}\})$ represents the residual mapping to be learned by three convolution layers. Each convolution layer is followed by a batch normalization (BN) layer and ReLU activation function.
Every convolution of the first residual graph block is 32 channels, the second is 64 channels. The receptive field is set to 6 (i.e., K=6). For simplicity, the residual graph CNN is named as G\underline{\hspace{0.5em}}ResNet, and the corresponding architecture is illustrated in Fig.\ref{res}.


\subsection{Inception Graph CNN}
\begin{figure}[t]
	\centering
	\subfigure[G\underline{\hspace{0.5em}}ResNet]{\includegraphics[scale=0.65]{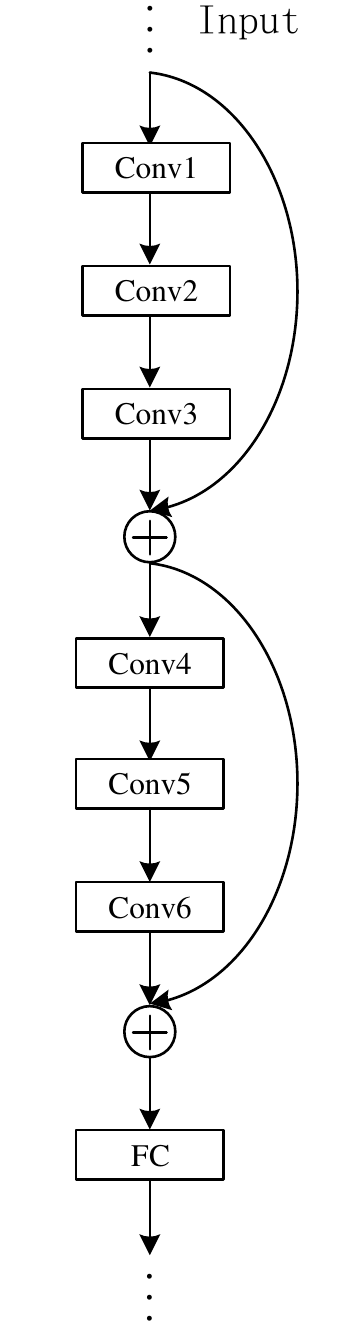} \label{res}}
	\subfigure[G\underline{\hspace{0.5em}}Inception]{\includegraphics[scale=0.65]{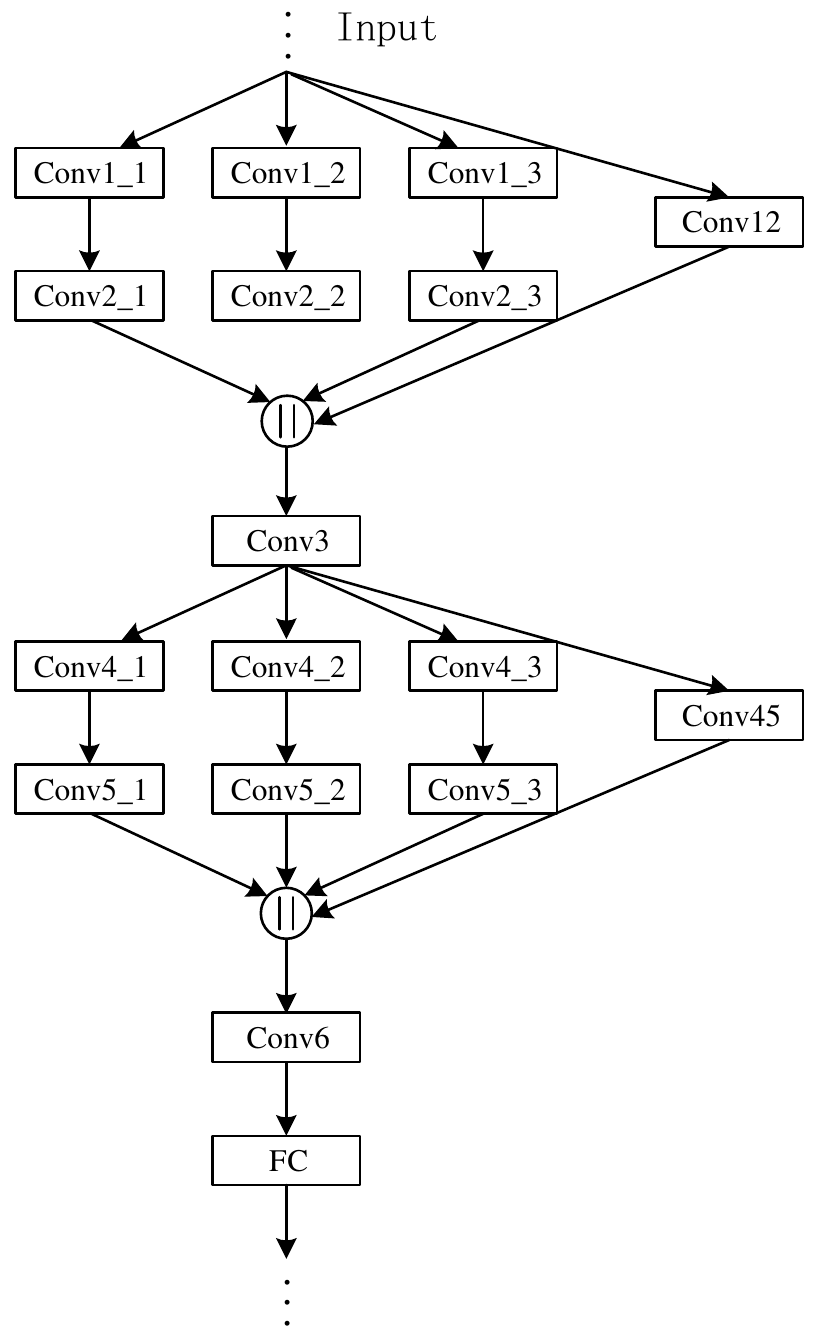} \label{incp}}
	\vspace{-1em}
	\caption{Graph CNN architectures of G\underline{\hspace{0.5em}}ResNet and G\underline{\hspace{0.5em}}Inception.}
	\label{sturcture2}
\end{figure}

For the Inception graph CNN named as G\underline{\hspace{0.5em}}Inception, we stack two Inception graph blocks that both followed by one convolution layer \ref{incp}.
%
Each Inception graph block is composed of four tributaries. Each tributary of the first three tributaries contains two convolutions and the last tributary contains one convolution.
According to Eq. \ref{cheby}, we choose convolution layers with different receptive fields at different tributaries for better capturing various information of graph data.
For the $j$-th convolution of the $i$-th tributary, $\mathbf{\Theta}_{ij}$ represents the corresponding parameter matrix. The Inception graph block can be given as
\begin{equation}
\begin{aligned}
\mathbf{Y} = & \mathcal{F}_{\text{inc}}(\mathbf{\bar{X}}_{K1}, \{\mathbf{\Theta}_{00}, \mathbf{\Theta}_{01}\}) \\
    & || \ \mathcal{F}_{\text{inc}}(\mathbf{\bar{X}}_{K2}, \{\mathbf{\Theta}_{10}, \mathbf{\Theta}_{11}\}) \\
    & || \ \mathcal{F}_{\text{inc}}(\mathbf{\bar{X}}_{K3}, \{\mathbf{\Theta}_{20}, \mathbf{\Theta}_{21}\}) \\
    & || \ \mathcal{F}_{\text{inc}}(\mathbf{\bar{X}}_{K4}, \{\mathbf{\Theta}_{30}\}), \\
\end{aligned}
\end{equation}
where the symbol $``||"$ represents concatenation \cite{mix} among convolution layers,
Each convolution layer is also followed by a batch normalization (BN) layer and ReLU activation function.
The function $\mathcal{F}_{\text{inc}}(\cdot)$ denotes how to represent the graph information with a prescribed receptive field in each path of inception block, and receptive fields are not exactly the same in different tributaries.

In the graph Inception block, the receptive fields of convolution layers in the first three paths can be set as $K_1=3$, $K_2=6$ and $K_3=9$, respectively. And the rest path only includes one convolution layer, and we choose $K_4=6$ as the size of receptive field.
Such a designed Inceptive structure can take multi-scale information of graph signal into account. The convolution layers in the first Inception graph block is with 32 channels, the second is with 64 channels. The receptive field is set to 6 of both ``Conv3" with 32 channels and ``Conv6" with 64 channels followed by two graph Inception block.

\section{Experiments}
\label{sec:exp}
In the section, we evaluate the performance of four different graph CNN structures on several public benchmark datasets including bioinformatics and social network datasets. They are all undirected graphs and their global properties are summarized in Table \ref{tab1}  \cite{DyGraph2017}. We first introduce the datasets and experimental setups, and then report and analyze the experimental results, after which a further discussion will be held about different parameters of Graph CNN.

\subsection{Datasets}
\textbf{Bioinformatics datasets}.
MUTAG \cite{mutag} is a nitro compounds dataset including 188 samples and divided into 2 classes.
PTC \cite{ptc} consists of compounds labeled according to carcinogenicity on rodents with 19 vertex labels.
NCI109 \cite{nci1andnci109} is a balanced dataset of chemical compounds screened for activity against non-small cell lung cancer and ovarian cancer cell, and they contain 4110 and 4127 chemical compounds, respectively.
ENZYMES \cite{enzymes} is a dataset of 600 protein tertiary structures obtained from the BRENDA enzyme database. The ENZYMES dataset contains 6 enzymes.

\textbf{Social network datasets}. These social network datasets come from \cite{deepgraphkerner}. We use the number of neighbors of each node as the label of the node.
COLLAB is a scientific collaboration dataset containing ego-networks of different researchers from three subfields of Physics. The task can then determine whether the ego-collaboration network belongs to any of three classes: High Energy Physics, Condense Mater Physics and Astro Physics.
IMDB-BINARY and IMDB-MULTI are movie collaboration datasets. Each graph represents a movie. Each vertex represents an actor that appears in the movie. IMDB-BINARY are constructed from Action and Romance genres. IMDB-MULTI contain three classes: Comedy, Romance and Sci-Fi. The task is to predict which genre a graph belongs to.

\begin{table*}[!t]
	\centering
	\renewcommand\arraystretch{1.3}
	\caption{Summary of graph datasets used in our experiments.}
	\footnotesize
	\begin{tabular}[width=0.8\linewidth]{l|ccccc} %
		\hline  Dataset & Num graphs &Classes  &Node labels &Avg nodes &Avg edges \\
		\hline
		MUTAG \cite{mutag}   &188  &2 &7  &17.93 &19.79\\
		PTC \cite{ptc}     &344  &2 &19 &14.29 &14.69 \\
		NCI109 \cite{nci1andnci109}   &4127 &2 &38 &29.68 &32.13\\
		ENZYMES \cite{enzymes}  &600  &6 &3  &32.63 &62.14\\
		\hline
		COLLAB \cite{deepgraphkerner}         &5000  &3  &-  &74.49   &2457.78\\
		IMDB-BINARY \cite{deepgraphkerner}     &1000  &2  &-  &19.77   &96.53  \\
		IMDB-MULTI \cite{deepgraphkerner}     &1500  &3  &-  &13.0    &65.94 \\
		\hline
	\end{tabular} \vspace{0.15cm}
	
	\label{tab1} \vspace{-0.1cm}
\end{table*}

\subsection{Experimental Setups}
We train four different graph CNNs, which structures has been described in Section IV. Each dataset is divided into 10 groups, of which nine are used to train the network and the remaining one is used for testing.
We carry out 10-fold cross validation, and its average results can be used as the final accuracy rates.
The prediction accuracy is expressed in form of $``\text{standard} \pm \text{deviation}"$ in graph classification benchmark datasets.
For each cross-validation, we train 300 epochs using Momentum Optimizer and the learning rate is set to 0.01. In addition, the momentum is 0.9 and the decay rate is with 0.95. We use cross entropy as loss function.
For the baseline Graph CNN, G\underline{\hspace{0.5em}}ResNet and G\underline{\hspace{0.5em}}DenseNet, the receptive field of each convolution layer is set to $6$.
For G\underline{\hspace{0.5em}}Inception, it can merge multi-scale information of graph signals. Therefore, in the G\underline{\hspace{0.5em}} Inception block, for each of the two convolutions, we set $3$, $6$ and $9$ as the receptive fields, respectively. And for the rest only one convolution, We choose the middle $6$ as its receptive field. To prevent overfitting, we add a dropout layer followed the fully connected layer, which can randomly discard half of the neurons.

\subsection{Results and comparisons}
We compare the proposed Graph CNNs (i.e., baseline Graph CNN, G\underline{\hspace{0.5em}}ResNet, G\underline{\hspace{0.5em}}DenseNet and G\underline{\hspace{0.5em}}Inception) with several state-of-the-art approaches on seven graph datasets, such as random walk kernel (RW) \cite{rw2003}, the graphlet kernels (GK) \cite{gk2009}, the Weisfeiler-Lehman subtree kernels (WL) \cite{wl2011}, Feature-Based (FB) \cite{BrunaZSL13}, Deep Graphlet (DGK) and DWL \cite{DGKandDWL2015}, the convolutional neural network (PSCN) \cite{niepert2016learning}, the shift aggregate extract network (SAEN) \cite{saen2017} and the dynamics based features (DyF) \cite{DyGraph2017}.
The performance of four Graph CNNs and comparisons with several state-of-the-art methods are shown in table \ref{compres}.
Overall, the experimental performance of the our four Graph CNN structures is superior to other existing methods.
Comparing with kernel based method \cite{rw2003,gk2009,wl2011,DGKandDWL2015}, feature based method \cite{BrunaZSL13,DyGraph2017}, convolution neural network (PSCN) \cite{niepert2016learning} and shift aggregate extract network (SAEN) \cite{saen2017}, all our graph CNN structures with spectral filtering method achieve the state-of-the-art performance on all datasets and obtain a significantly improvement on MUTAG, PTC, ENZYMES and IMDB-BINARY datasets in contrast to the second best performance.
G\underline{\hspace{0.5em}}ResNet achieves state-of-the-art on two social network datasets: $79.90\%$ vs $72.87\%$ \cite{DyGraph2017} on IMDB-BINARY and $54.43\%$ vs $50.55\%$ \cite{wl2011} on IMDB-MULTI.
G\underline{\hspace{0.5em}}Inception can significantly outperform these state-of-the-art algorithms on two datasets: $95.00\%$ vs $92.63\%$ \cite{niepert2016learning} on MUTAG and $67.50\%$ vs $53.43\%$ \cite{DGKandDWL2015} on ENZYMES.
G\underline{\hspace{0.5em}}DenseNet shows much improvement than other methods on two  bioinformatic datasets: $73.24\%$ vs $60.00\%$ \cite{niepert2016learning} on PTC and $80.66\%$ vs $80.32\%$ \cite{DGKandDWL2015} on NCI109 and $83.16\%$ vs $80.61\%$ \cite{DyGraph2017} on COLLAB.
Our graph CNNs are able to render very impressive results. The performance pf graph recognition can be further improved to some extent when we deepen these three graph CNN and the results are reported in the following discussion.

When comparing the performance of the baseline Graph CNN, G\underline{\hspace{0.5em}}ResNet, G\underline{\hspace{0.5em}}Inception and G\underline{\hspace{0.5em}}DenseNet, different Graph CNN models have the certain advantages on various types of graph datasets.
The structures of ResNet, Inception and DenseNet were proposed initially to boost the performance of deep convolution networks from different aspects, such as residual learning, considering information in multiple receptive fields, densely employing multi-level representations.
For better show how to transform these CNN structures to Graph CNNs, the proposed G\underline{\hspace{0.5em}}ResNet, G\underline{\hspace{0.5em}}Inception and G\underline{\hspace{0.5em}}DenseNet can still outperform our baseline network, which is a simple 6-layer graph CNN framework.
For example, our  G\underline{\hspace{0.5em}}Inception model on the ENZYMES dataset can significantly outperform the baseline network, 67.50\% vs 64.83\%. 	This indicates that our  G\underline{\hspace{0.5em}}Inception model can introduce greater data diversity in multiple receptive fields and improve the network performance of graph data.
We achieve much better performance with the G\underline{\hspace{0.5em}}DenseNet framework then the baseline network on PTC dataset, e.g., 73.24\% vs 71.76\%. It demonstrates that the G\underline{\hspace{0.5em}}DenseNet perform very well on the graph recognition problem by simultaneously considering different level information.
The accuracies of G\underline{\hspace{0.5em}}ResNet have been improved on IMDB-BINARY and IMDB-MULTI datasets, which also show its capability by transforming the ResNet structure to Graph CNN.




\begin{table*}[!t]
	\centering
	\renewcommand\arraystretch{1.3}
	\caption{Comparison of graph recognition performances with different Graph CNN models and several state-of-the-arts on graph datasets.}
	\footnotesize
	\begin{tabular}[width=0.8\linewidth]{l|ccccccc} %
		\hline  Dataset & MUTAG &PTC  &NCI109 &ENZYMES &COLLAB &IMDB-B &IMDB-M\\
		\hline	\hline
		RW \cite{rw2003}   &$83.72\pm1.50$ &$57.85\pm1.30$  &$49.75\pm0.60$  &$24.16\pm1.64$  &$69.01\pm0.09$   &$64.54\pm1.22$   &$34.54\pm0.76$\\
		GK \cite{gk2009}  &$81.66\pm2.11$ &$57.26\pm1.41$  &$62.60\pm0.19$  &$26.61\pm0.99$   &$72.84\pm0.28$   &$65.87\pm0.98$   &$43.89\pm0.38$\\
		WL \cite{wl2011}  &$80.72\pm3.00$ &$56.97\pm2.01$  &$80.22\pm0.34$  &$53.15\pm1.14$   &$77.79\pm0.19$   &$72.86\pm0.76$   &$50.55\pm0.55$\\
		FB \cite{BrunaZSL13}  &$84.66\pm2.01$ &$55.58\pm2.30$  &$62.43\pm1.13$  &$29.00\pm1.16$   &$76.35\pm1.64$   &$72.02\pm4.71$   &$47.34\pm3.56$\\
		DGK \cite{DGKandDWL2015}  &$82.66\pm1.45$ &$57.32\pm1.13$  &$62.69\pm0.23$  &$27.08\pm0.79$   &$73.09\pm0.25$   &$66.96\pm0.56$   &$44.55\pm0.52$\\
		DWL \cite{DGKandDWL2015}  &$82.94\pm2.68$ &$59.17\pm1.56$  &$80.32\pm0.33$  &$53.43\pm0.91$   &-  &-  &-\\
		PSCN \cite{niepert2016learning} &$92.63\pm4.21$ &$60.00\pm4.82$   &-              &-                &$72.60\pm2.15$   &$71.00\pm2.29$  &$45.23\pm2.84$\\
		SAEN \cite{saen2017} &$84.99\pm1.82$ &$57.04\pm1.30$   &-              &-                &$75.63\pm0.31$   &$71.26\pm0.74$  &$49.11\pm0.64$\\
		DyF \cite{DyGraph2017}  &$88.00\pm2.37$ &$57.15\pm1.47$  &$66.72\pm0.20$ &$33.21\pm1.20$  &$80.61\pm1.60$   &$72.87\pm4.05$  &$48.12\pm3.56$\\
		\hline \hline
		Our baseline   &$93.89\pm6.31$  &$71.76\pm7.58$  &$80.51\pm2.67$  &$64.83\pm5.45$  &$82.96\pm0.86$  &$79.70\pm3.66$  &$54.40\pm4.88$\\		
		G\underline{\hspace{0.5em}}ResNet  &$94.44\pm5.56$ &$73.24\pm8.05$ &$80.27\pm2.56$ &$66.83\pm7.47$  &$82.64\pm0.99$  &\bm{$79.90\pm3.96$} &\bm{$54.53\pm4.25$}\\
		G\underline{\hspace{0.5em}}Inception  & \bm{$95.00\pm4.61$}  &$72.94\pm6.28$   &$80.32\pm1.73$  &\bm{$67.50\pm5.54$}  &$82.58\pm1.28$  &$78.40\pm3.72$  &$54.53\pm4.71$\\
		G\underline{\hspace{0.5em}}DenseNet   &$94.44\pm4.30$  &\bm{$73.24\pm6.64$}    &\bm{$80.66\pm2.49$}  &$66.83\pm4.86$  &\bm{$83.16\pm1.00$}  &$79.20\pm4.19$  &$54.40\pm4.70$ \\
		\hline
	\end{tabular} \vspace{0.15cm}
	\label{compres} \vspace{-0.1cm}
\end{table*}

\begin{figure*}[ht]
	\centering
	\subfigure[G\underline{\hspace{0.5em}}ResNet]{\includegraphics[scale=0.43]{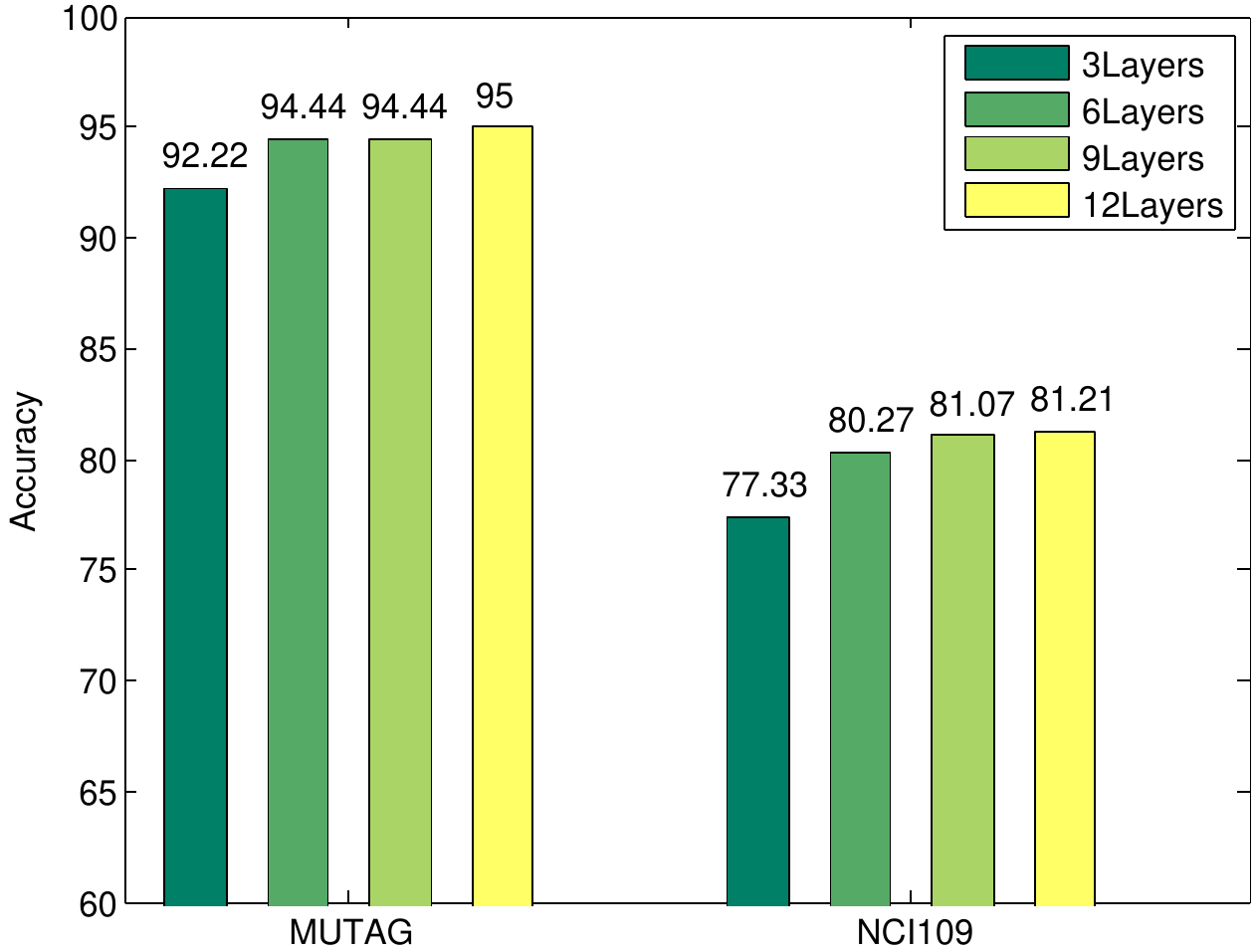} \label{layes_res}}
	\subfigure[G\underline{\hspace{0.5em}}Inception]{\includegraphics[scale=0.43]{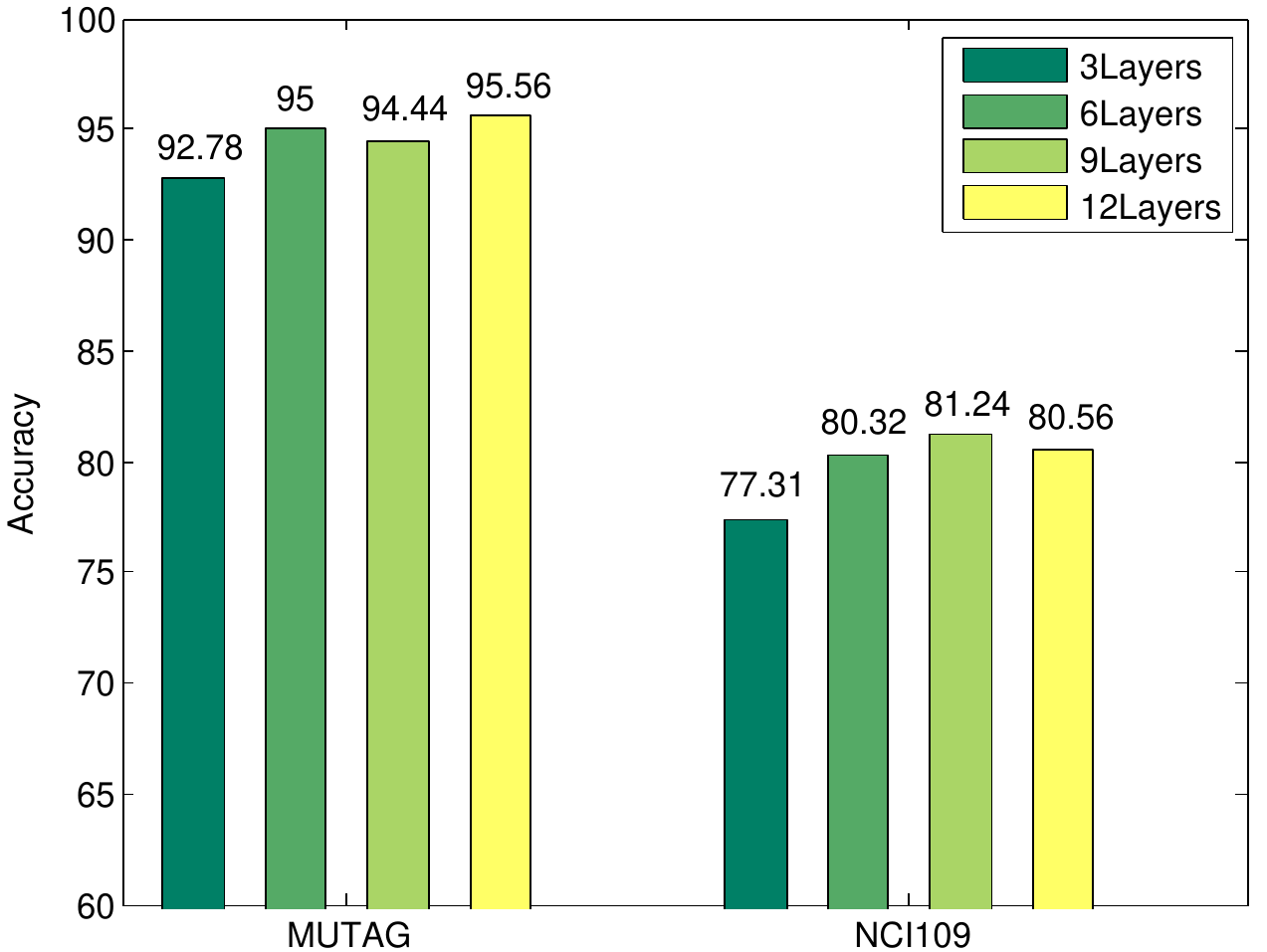} \label{layes_inp} }
	\subfigure[G\underline{\hspace{0.5em}}Densenet]{\includegraphics[scale=0.43]{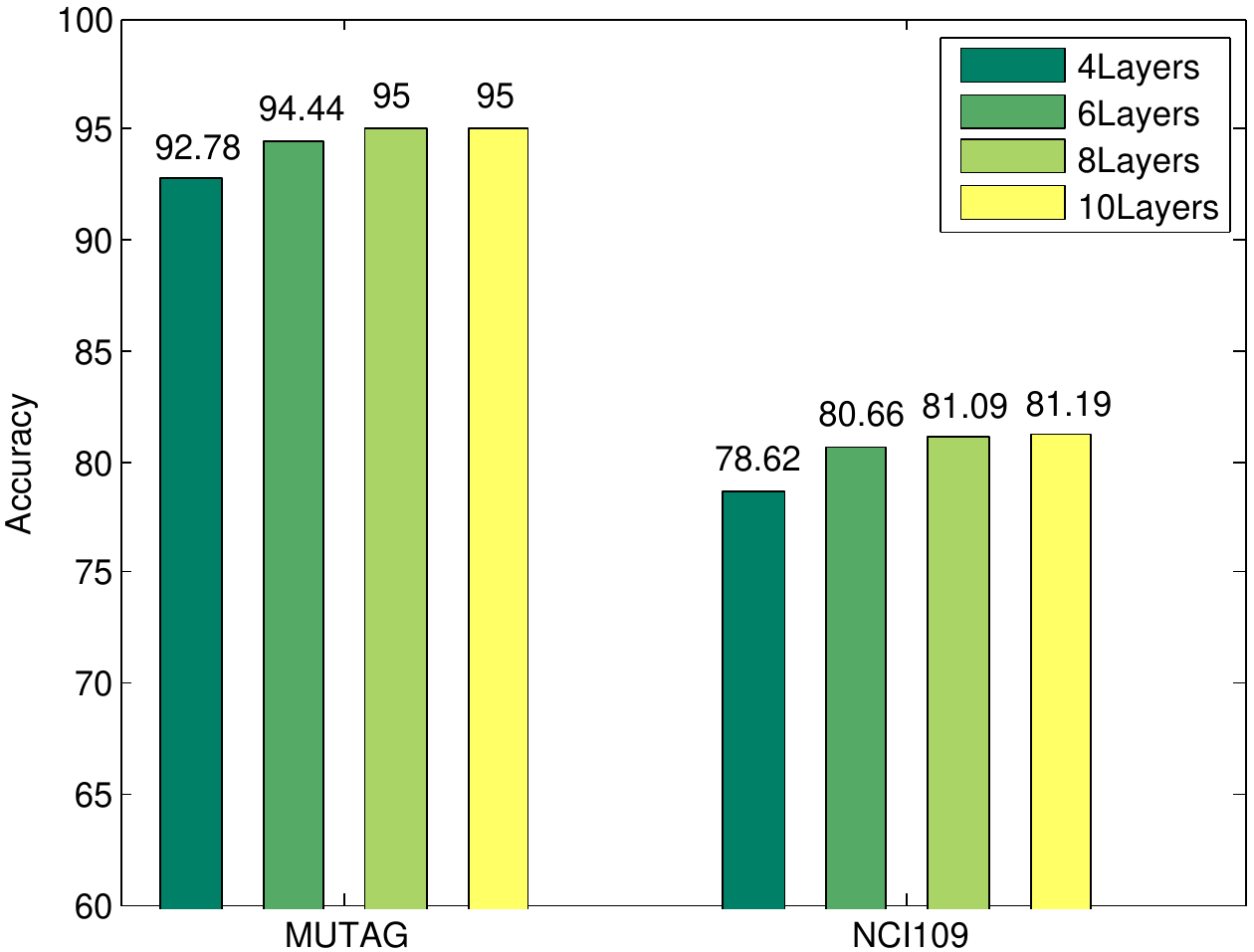} \label{layes_dens} }
	\vspace{-1em}
	\caption{Performance comparison of G\underline{\hspace{0.5em}}ResNet, G\underline{\hspace{0.5em}}Inception, G\underline{\hspace{0.5em}}DenseNet with different number of convolution layers. Experiments are conducted on two datasets : MUTAG (a small dataset with 188 nodes) and NCI109 (a large dataset with 4127 nodes). }
	\label{comparisonLayers}
\end{figure*}

\begin{figure*}[ht]
	\centering
	\subfigure[G\underline{\hspace{0.5em}}ResNet]{\includegraphics[scale=0.6]{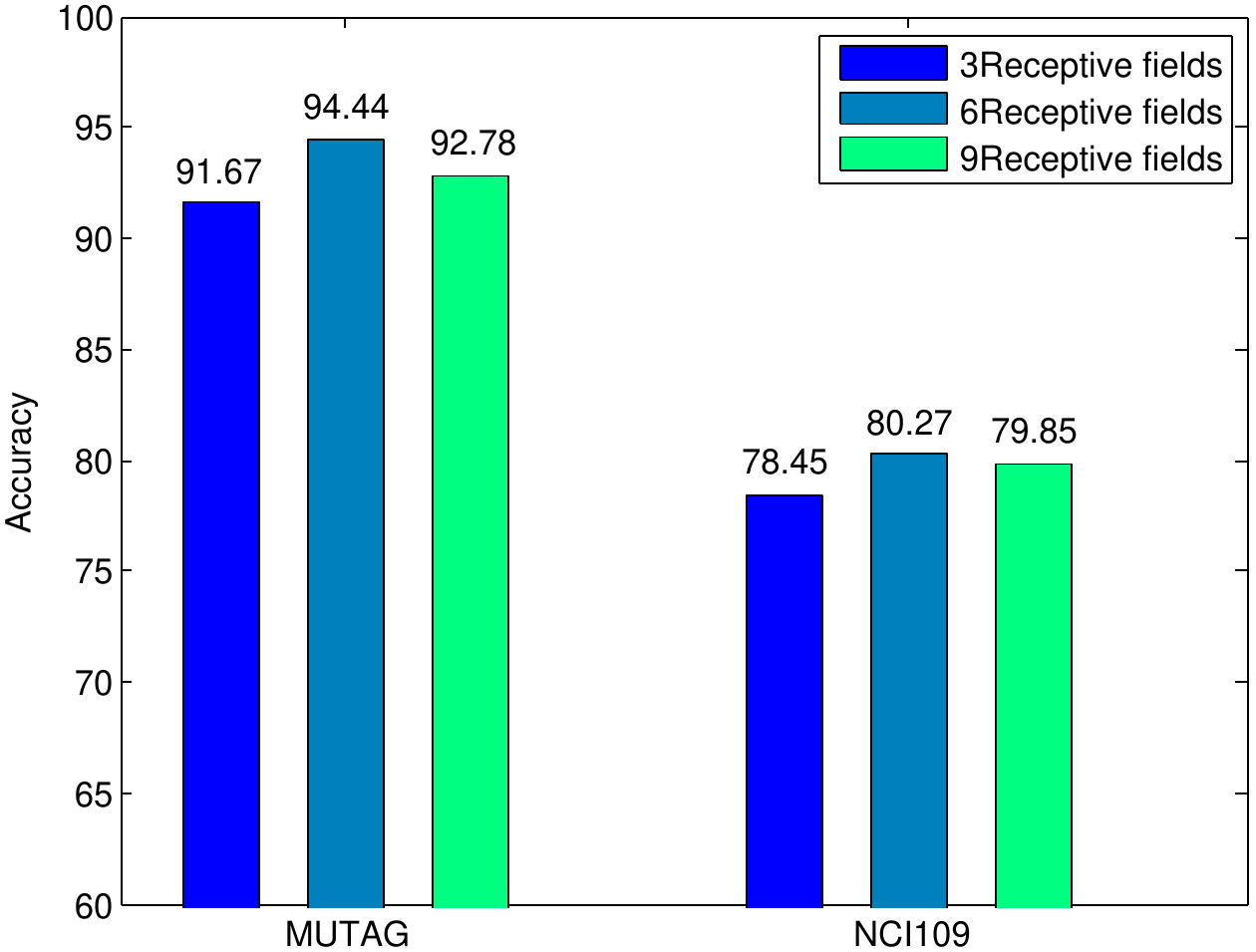} \label{ks_res}}
	\subfigure[G\underline{\hspace{0.5em}}Densenet]{\includegraphics[scale=0.6]{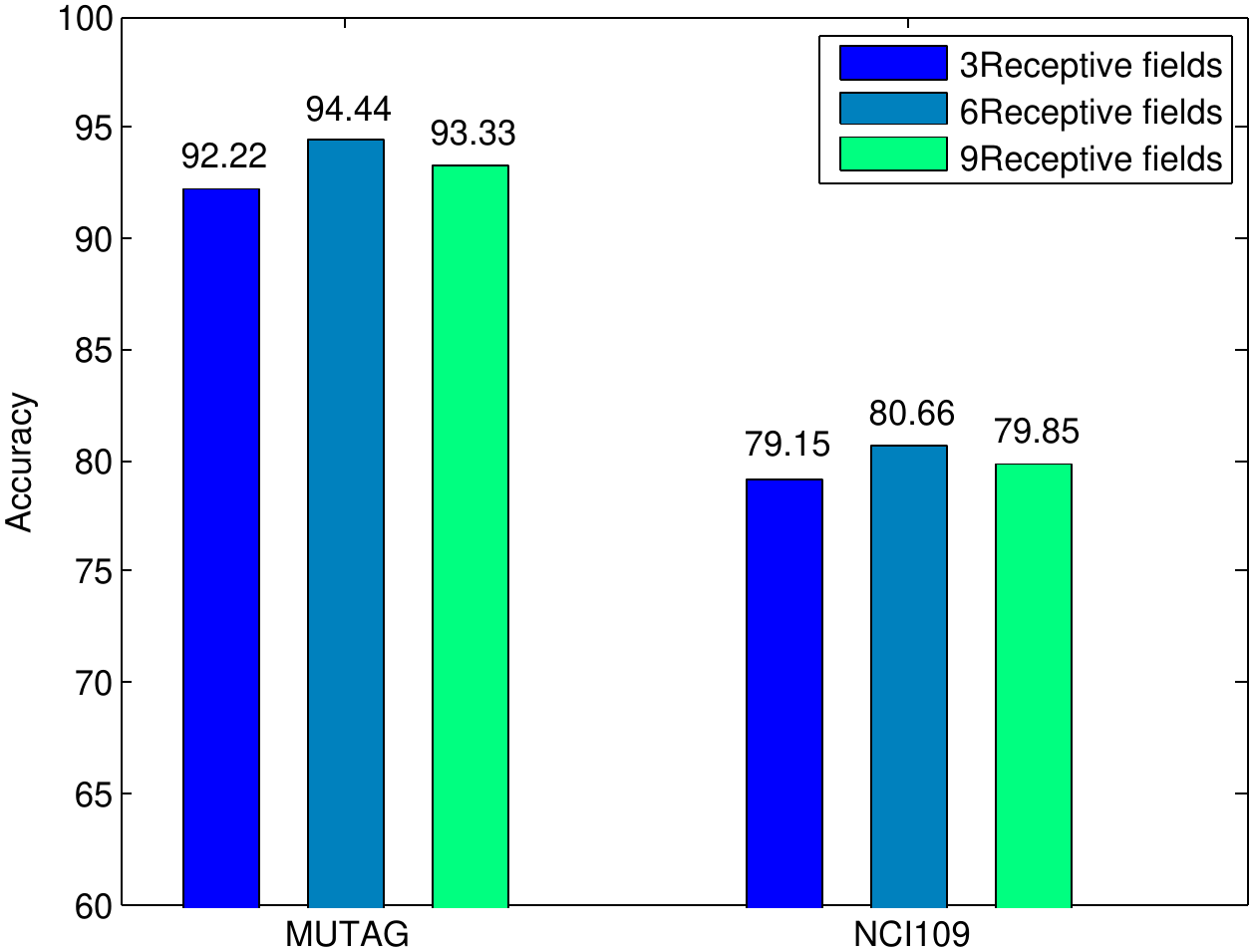} \label{ks_dens} \vspace{10em}}
	\vspace{-1em}
	\caption{Performance comparison of G\underline{\hspace{0.5em}}ResNet and G\underline{\hspace{0.5em}}Densenet with different receptive fields on two datasets : MUTAG (a small dataset with 188 nodes) and NCI109 (a large dataset with 4127 nodes). The size of receptive field can be set to 3, 6 and 9, respectively.}
	\label{comparisonkernelsize}
\end{figure*}

\subsection{Discussion}
\textbf{Deeper graph CNN:}
Here we explore how deeper graph CNNs impact the performance of graph recognition.
For G\underline{\hspace{0.5em}}ResNet and G\underline{\hspace{0.5em}}Inception networks, the number of convolution layers can be set to: 3 layers, 6 layers, 9 layers and 12 layers, respectively.
Considering that each previous result in G\underline{\hspace{0.5em}}DenseNet will be used as the input to the convolution layer behind. In this way, the number of parameters in the last few layers is huge, and the amount of calculations also increases dramatically. Therefore the number of convolution layers is set to: 4 layers, 6 layers, 8 layers and 10 layers for G\underline{\hspace{0.5em}}DenseNet.
We choose a small dataset MUTAG with 188 nodes and a large dataset NCI109 with 4127 nodes described above. All experiments of exploring deeper graph CNN are performed on these two datasets, and the experimental results are reported in Fig. \ref{comparisonLayers}.

For the G\underline{\hspace{0.5em}}ResNet model, we stack 3-layer convolution for each residual graph block.
In order to build deeper G\underline{\hspace{0.5em}}ResNet, we stack 6-layer, 9-layer, 12-layer convolution by 2, 3, 4 residual graph block, respectively.
The Fig. \ref{layes_res} shows the comparisons of different G\underline{\hspace{0.5em}}ResNet with different number of network layers on MUTAG and NCI109 datasets.
For MUTAG dataset, the results of G\underline{\hspace{0.5em}}ResNet models, which are with  3-layer convolution, the 6-layer convolution, the 9-layer convolution, and the 12-layer convolution, can achieve $92.22\%$, $94.44\%$, $94.44\%$ and $95\%$, respectively.
For NCI109 dataset, the performance of different G\underline{\hspace{0.5em}}ResNet models are $77.33\%$, $80.27\%$, $81.07\%$ and $81.21\%$, respectively.
%
In the G\underline{\hspace{0.5em}}Inception network, we further observe the changes of classification performance by stacking different number of inception blocks. Stacking 3-layer convolution means that there is only one Inception graph block in the G\underline{\hspace{0.5em}}Inception model.
Stacking 6-layer, 9-layer, 12-layer convolution means that there are 2,3,4 Inception graph block in G\underline{\hspace{0.5em}}Inception, respectively.
As can be illustrated in the Fig. \ref{layes_inp}, the performance of 3-layer G\underline{\hspace{0.5em}}Inception model is $92.78\%$ on MUTAG dataset.
The G\underline{\hspace{0.5em}}Inception with 12-layer convolution is better than with the 3-layer convolution, e.g., 95.56\% vs 92.78\%.
For NCI109 dataset, the results with different deeper G\underline{\hspace{0.5em}}Inception models are $77.31\%$, $80.32\%$, $81.24\%$ and $80.56\%$, respectively. The G\underline{\hspace{0.5em}}Inception network with 9-layer convolution achieves the highest performance $81.24\%$ and improves the average accuracy by  $3.93\%$ than the G\underline{\hspace{0.5em}}Inception model with the 3-layer convolution.
%
%
Similar to comparisons of G\underline{\hspace{0.5em}}ResNet and G\underline{\hspace{0.5em}}Inception, we explore the performance of G\underline{\hspace{0.5em}}DenseNet with 4-layer, 6-layer, 8-layer and 10 layer network, respectively.
As reported in Fig. \ref{layes_dens}, the G\underline{\hspace{0.5em}}Inception network with the 12-layer convolution can outperform the G\underline{\hspace{0.5em}}Inception with 3-layer convolution by $2.22\%$.
On the NCI109 dataset, The G\underline{\hspace{0.5em}}Inception with 12-layer convolution is better than with the 3-layer convolution, e.g., 81.19\% vs 78.62\%.
We can observe that as the number of convolution layers increases, the classification accuracy increases. Another reasonable explanation is that deeper Graph CNN models abstract higher-level presentations that will also help improve the performance of graph recognition.

\textbf{Extend the receptive field of convolution layer.} For G\underline{\hspace{0.5em}}ResNet and G\underline{\hspace{0.5em}}DenseNet, we also explore the performance of different receptive fields.
The structure of G\underline{\hspace{0.5em}}Inception has been combined with a variety of receptive fields information, so we don't conduct this discussion on it.
In different G\underline{\hspace{0.5em}}Inception networks, the size of receptive field is set to: 3, 6  and 9, respectively.
We think that the selection of receptive fields, which is important for graph CNN, can consider different local structural information.
The experiment is performed on the same two datasets: MUTAG and NCI109, and the comparisons of different receptive fields are reported in Fig. \ref{comparisonkernelsize}.
%
With the 3, 6 and 9 receptive fields of G\underline{\hspace{0.5em}}ResNet, we can observe accuracies of $91.67\%$, $94.44\%$, $92.78\%$ on MUTAG dataset, and $78.45\%$, $80.27\%$, $79.85\%$ on NCI109 dataset (see Fig. \ref{ks_res}).
When the receptive field is set to 6, the performance of G\underline{\hspace{0.5em}}ResNet is  highest on both datasets.
As can be reported in Fig. \ref{ks_dens}, we can observe that different G\underline{\hspace{0.5em}}DenseNet models with 3, 6, 9 receptive fields can gain
$92.22\%$, $94.44\%$, $93.33\%$ on MUTAG dataset, and $79.15\%$, $80.66\%$, $79.85\%$ on NCI109 dataset.
The performance is also highest when the receptive field is set to 6 in the G\underline{\hspace{0.5em}}DenseNet model.
It demonstrates that extracted representations of graph signals are overly complex and too small to represent useful features, when the size of the receptive field is too large.
If the receptive field is too small, local structural features may not be extracted, and the performance of graph CNN will be reduced.


\section{Conclusions and Future Work}
\label{sec:con}
In this work, we have given a comprehensive analysis of when work matters by transforming classical network structures to graph CNN, particularly for the basic graph recognition task.
Inspired by the basic ideas of ResNet, DenseNet and Inception network, we have constructed different Graph CNN architectures, including the plain graph CNN, G\underline{\hspace{0.5em}}ResNet, G\underline{\hspace{0.5em}}Inception, G\underline{\hspace{0.5em}}DenseNet.
By constructing the G\underline{\hspace{0.5em}}Inception network, we focus on considering different receptive fields of graph signals in the process of convolution, while G\underline{\hspace{0.5em}}DenseNet is responsible for capturing different-level representations of graph CNN.
The G\underline{\hspace{0.5em}}ResNet is benefit to constructing deeper graph networks with the help of residual learning.
Extensive experimental results clearly demonstrated that effective of the proposed G\underline{\hspace{0.5em}}ResNet, G\underline{\hspace{0.5em}}Inception, G\underline{\hspace{0.5em}}DenseNet models for the problem of graph recognition.
In the future, we will further extent the Graph CNN architectures for generic understanding tasks, e.g., image restoration, social network analysis, visual understanding, etc.



\section*{Acknowledgment}
This work was supported by the National Natural Science Foundation of China (Grant 61073094 and U1233119). This research was partly supported under Australian Research Council Discovery Projects funding scheme (project DP140102270).


	\bibliographystyle{IEEEtran}
	\bibliography{mybibfile}

\end{document}